\documentclass{article}
\usepackage{spconf}
\usepackage{cite}
\usepackage{amsmath,amssymb,amsfonts}
\usepackage{algorithmic}
\usepackage{graphicx}
\usepackage{textcomp}
\usepackage{multirow}
\usepackage{xcolor}
\usepackage{gensymb}
\usepackage{subfig} 
\usepackage{float} 

\title{FEW-SHOT CLASS-INCREMENTAL LEARNING FOR EFFICIENT SAR AUTOMATIC TARGET RECOGNITION}

\name{George Karantaidis\qquad Athanasios Pantsios \qquad Ioannis Kompatsiaris \qquad Symeon Papadopoulos \thanks{This project has received funding from the FaRADAI project (ref. 101103386) funded by the European Commission under the European Defence Fund. Views and opinions expressed are however those of the author(s) only and do not necessarily reflect those of the European Union or the European Commission. Neither the European Union nor the granting authority can be held responsible for them.}}
\address{Information Technologies Institute, Centre for Research \& Technology Hellas, Thessaloniki, Greece.}

\begin{document}
\maketitle

\begin{abstract}

Synthetic aperture radar automatic target recognition (SAR-ATR) systems have rapidly evolved to tackle incremental recognition challenges in operational settings. Data scarcity remains a major hurdle that conventional SAR-ATR techniques struggle to address. To cope with this challenge, we propose a few-shot class-incremental learning (FSCIL) framework based on a dual-branch architecture that focuses on local feature extraction and leverages the discrete Fourier transform and global filters to capture long-term spatial dependencies. This incorporates a lightweight cross-attention mechanism that fuses domain-specific features with  global dependencies to ensure robust feature interaction, while maintaining computational efficiency by introducing minimal scale-shift parameters. The framework combines focal loss for class distinction under  imbalance and center loss for compact intra-class distributions to enhance class separation boundaries. Experimental results on the MSTAR benchmark dataset demonstrate that the proposed framework consistently outperforms state-of-the-art methods in FSCIL SAR-ATR, attesting to its effectiveness in real-world scenarios.
\end{abstract}


\begin{figure} 
\centering
\includegraphics[width=\columnwidth]{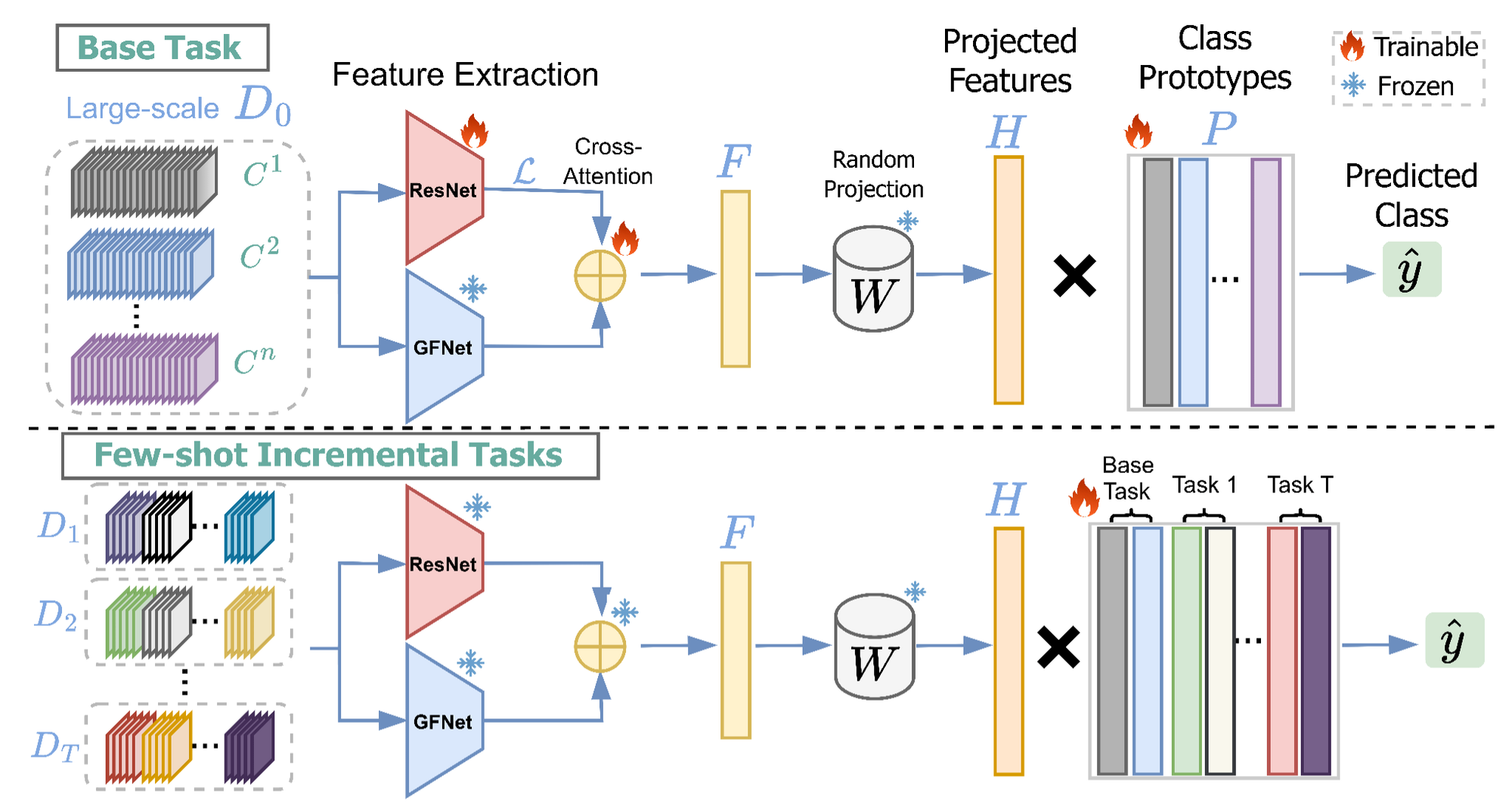}
\caption{Overview of the proposed DILHyFS framework for FSCIL SAR-ATR.}
\label{fig:workflow}
\end{figure}

\section{Introduction}

Synthetic Aperture Radar (SAR) technology demonstrates superior capabilities in diverse and demanding  operational conditions, and is widely deployed in applications, such as environmental monitoring, disaster management, and defense. With advances in deep learning and memory-efficient approaches, significant efforts have been directed toward advancing the challenging task of SAR Automatic Target Recognition (ATR).

SAR-ATR algorithms demonstrate exceptional performance in practical scenarios and applications. However, their static nature limits adaptability in practical settings. Learning incrementally \cite{rebuffi2017icarl, hou2019learning} constitutes a significant challenge, as operational conditions require  algorithms to continuously adapt to new classes encountered in dynamic environments, while retaining prior knowledge. These approaches tackle ``catastrophic forgetting'' \cite{song2024overcoming}, which degrades performance in streamed-data scenarios. Challenges in SAR-ATR arise mainly from intra-class variation and inter-class similarity. Unlike natural images, SAR target discrimination relies on sparse backscatter patterns. Inter-class confusion occurs when similar targets at certain azimuth angles produce overlapping patterns, while intra-class variation stems from changes in azimuth and depression angles of the same target.

In several cases, SAR-ATR systems face challenges due to limited labeled data calling for data-efficient algorithms. Conventional methods that store  and reuse all past data are impractical due to high memory and processing requirements. Few-shot class-incremental learning (FSCIL) approaches address these challenges by enabling adaptation to new tasks with minimal labeled data, while  avoiding the storage of extensive past ones. 

In this paper, we propose a framework specifically designed for FSCIL in SAR-ATR. The proposed Dual-network Incremental Learning with Hybrid loss for Few-Shot SAR-ATR, called DILHyFS, integrates a dual-branch architecture, combining ResNet-18 \cite{he2016deep} and GFNet \cite{rao2021global} to address SAR-ATR challenges, as demonstrated in Fig.~\ref{fig:workflow}. GFNet is a computationally efficient architecture that learns long-range spatial dependencies in the frequency domain using a discrete Fourier transform and global filters. This enables capturing complex interactions in SAR spectrograms, enhancing feature extraction with limited labeled data. Additionally, a cross-attention mechanism is employed to associate the outputs of the subsequent layers of both GFNet and ResNet-18. This fusion technique enables the frozen GFNet layers to guide the training of ResNet-18 layers, resulting in robust feature representations that integrate the domain-specific features learned by ResNet-18 with the long-range spatial dependencies captured by GFNet. Furthermore, this is a lightweight fusion method that enhances feature representation by introducing only a small number of scale-shift parameters between layers. To further optimize performance, we introduce a novel loss function that combines focal and center loss. The focal loss improves class distinction, particularly under class imbalance, while the center loss promotes compact intra-class distributions. 
Extensive experiments across various setups, emulating real-world scenarios on the MSTAR dataset, demonstrate the effectiveness of our method, exhibiting significant improvements over state-of-the-art approaches, highlighting the potential of DILHyFS for efficient FSCIL in SAR-ATR applications. Our contributions are summarized as follows:

\begin{itemize}
    \item We propose DILHyFS, a dual-branch architecture for FSCIL that combines feature representations from ResNet-18 with long-range spatial dependencies captured by GFNet to address limited labeled data and incremental learning challenges in SAR-ATR.
    \item We introduce a lightweight cross-attention mechanism to align and fuse the extracted features  to enhance feature representation and improve the model's robustness and generalization for incremental learning tasks. 
    \item Our framework introduces a hybrid loss combining focal loss, which enables class separation, and center loss, which promotes compact intra-class distributions and better class separation for incremental learning tasks.
    \item We validate the proposed framework on the MSTAR dataset, demonstrating significant improvements over state-of-the-art methods and its effectiveness  across various emulated FSCIL SAR-ATR scenarios. 
\end{itemize}

\section{Related work}
 
\subsection{Few-Shot Class-Incremental Learning}

FSCIL approaches can be broadly categorized into three groups: data-based, structure-based, and optimization-based methods. Data-based approaches address challenges like data scarcity and computational constraints by reusing or managing limited data effectively \cite{zhou2022forward, zhang2021few}. Structure-based methods focus on designing adaptable models, employing techniques like dynamic structures and attention mechanisms to integrate new classes while retaining prior knowledge \cite{tao2020few}. Optimization-based approaches tackle FSCIL challenges through strategies such as representation learning, meta-learning, and knowledge distillation, enhancing the model's ability to learn efficiently and retain knowledge \cite{hersche2022constrained, kim2023warping, kalla2022s3c}.

\subsection{SAR Target Classification}

Conventional SAR-ATR approaches  have shown remarkable performance and  have been extensively studied in controlled static scenarios with a large amount of labeled data, but often fail in settings where labeled data is scarce. This highlights the need for FSCIL approaches to effectively address data scarcity in dynamically changing situations. An azimuth-aware subspace classifier was proposed in \cite{zhao2024azimuth}  that benefits from Grassmannian manifold and leverages the azimuth dependence characteristics of SAR images. Also, a set of loss functions were proposed to balance stability and plasticity while addressing domain-specific challenges. A different strategy was introduced in \cite{zhao2023few}, where a cosine prototype learning framework was developed to address the challenge of balancing plasticity and stability. Losses were proposed to improve inter-class separability and intra-class compactness, allowing the model to learn new concepts without sacrificing its ability to generalize. A self-supervised classifier was proposed in \cite{zhao2024decoupled} to enable efficient knowledge transfer and stable discrimination across classes, where two self-supervised tasks were designed to generate virtual samples and enhance the discriminability of scattering patterns. In contrast, \cite{zhao2024few} introduced an augmentation module that synthesizes virtual targets with diverse scattering patterns to improve the model's forward compatibility for incremental classes. Meanwhile, \cite{kong2024few} proposed a random augmentation technique to improve generalization and strengthen class boundaries. Additionally, they employed an orthogonal distribution optimization strategy to reduce feature confusion and reserve space for unseen classes. FSCIL for SAR-ATR is in its early stages, with state-of-the-art methods achieving relatively low accuracies. Several challenges remain, emphasizing the need for further advancements.

\section{Method}
\subsection{Preliminaries}

FSCIL approaches are trained on a sequential dataset $D = \{D_0, D_1, \ldots, D_T\}$, where $D_0$ represents the base task and $D_t= \left\{ \left( x_i, y_i \right) \right\}_{i=1}^{|D_t|}$ the dataset for the $t^{th}$ incremental task with $t \in\{1, 2, \ldots, T\}$. The model is initially trained on $D_0$, which includes a large label space $C_0$ and sufficient data for each class. Then, the system encounters few-shot tasks $D_t$, consisting of  $|D_t| =k \cdot |C_t|$ samples and the model has access only to current task data. A FSCIL algorithm updates the model, retaining knowledge from previous tasks while learning new classes \cite{chen2021incremental}.

\subsection{Proposed Framework}
The proposed framework, DILHyFS, extends FSCIL for SAR-ATR by integrating a GFNet-based feature extractor, a cross-attention mechanism, and an improved loss function. These enhancements improve feature discrimination, mitigate catastrophic forgetting, and enhance generalization. Given a SAR image, the model extracts features using a dual-branch pipeline with ResNet-18 and GFNet. The cross-attention mechanism described in Sec. \ref{sec:cross_attention} fuses these features, enhancing the model’s ability to learn robust and discriminative representations.

The features $\mathbf{F}$ are extracted from the input SAR images and the procedure is repeated for each incremental task $t$.   To improve class separability, these features are projected onto a randomly initialized matrix $\mathbf{W}$ of dimension $M$, which remains frozen throughout training: $\mathbf{H} = \phi(\mathbf{F}^{\top} \mathbf{W})$. This projection enhances linear separability, ensuring that newly learned classes do not interfere with previously learned ones. In the next step, the mean prototypes $C$ are calculated as: $\mathbf{C} = \sum_{t=1}^{T}  \sum_{n=1}^{|D_t|} \tilde{\mathbf{H}}_{t,n} \otimes \mathbf{y}_{t,n}$, where $ \tilde{\mathbf{H}}_{t,n} = \mathbf{H}_{t,n} / m_{c} $ represents the normalized feature vector and ${m_{c}} $ the number of samples in the corresponding class. 
 
To further reduce intra-class variance and increase inter-class discrimination, an incremental LDA algorithm \cite{pang2005incremental},  is applied, leveraging the Gram matrix $\mathbf{G}$ of the projected features: $\mathbf{G} = \sum_{t=1}^{T} \sum_{n=1}^{|D_t|} \mathbf{H}_{t,n}\otimes \mathbf{H}_{t,n}$. Class prototypes $P$ are computed as $\mathbf{P = (G + \lambda I)^{-1} C}$, where $\lambda$ is the ridge regression parameter,  determined during the base task when sufficient data is available. The optimal value of $\lambda$ is selected  by minimizing the mean squared error between predictions and targets on a validation set ($20\%$ of the training data) and remains fixed thereafter. At inference, predictions for test samples are made using the learned class prototypes by $\hat{y} = \arg \max_{c \in \mathcal{Y}_t} ( \mathbf{H}_{test} \mathbf{P} )$, where $\mathbf{H}_{test}$ represents the extracted features of the test samples. Prototype-based approaches have shown promise in SAR-ATR \cite{karantaidis2025incsar}, and DILHyFS builds upon this foundation while introducing a dual-branch network, cross-attention fusion, and a hybrid loss function to further improve adaptability and scalability.

\subsection{Cross-Attention Mechanism} \label{sec:cross_attention}
Having extracted features from both branches, we employ a cross-attention mechanism that integrates two different backbones: ResNet and GFNet. Specifically, we employ ResNet-18, pre-trained on ImageNet-2012, and GFNet, pre-trained on ImageNet. 
A GFNet block transforms input spatial features into the frequency domain using a 2D Discrete Fourier Transform, $\mathcal{F}$. Learnable global filters $K$ are then applied, followed by a 2D Inverse Fourier Transform, $\mathcal{F}^{-1}$, to map the features back to the spatial domain: 
\begin{equation}
x \leftarrow \mathcal{F}^{-1} \left[ K \odot \mathcal{F}[x] \right]    
\end{equation}

Global filters help GFNet blocks capture distinct patterns in the frequency domain, allowing them to focus on frequency-domain relationships rather than spatial-domain \cite{rao2021global}. 
Finally, a feed forward network with a normalization layer and a two-layer MLP network is applied. GFNet consists of four layers that have $3$, $3$, $10$, $3$ blocks, respectively.

\begin{figure} 
\centering
\includegraphics[width=0.95\linewidth]{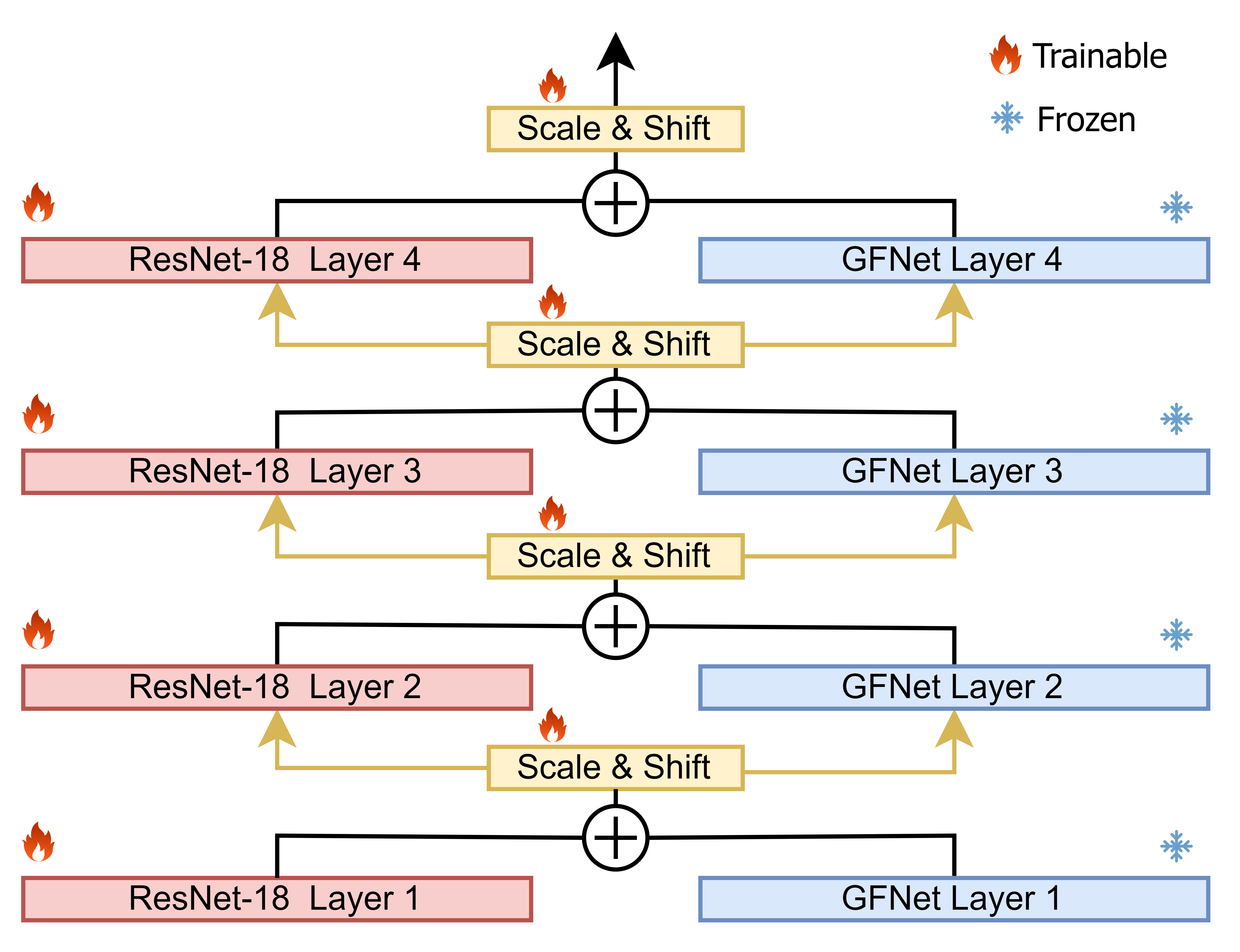}
\caption{Illustration of the proposed feature fusion cross-attention module.}
\label{fig:Diagram}
\end{figure}

A cross-attention module is proposed for the fusion of the two networks. Both models have a hierarchical structure with four layers with $64$, $128$, $256$, $512$ dimensions, respectively. As depicted in Fig.~\ref{fig:Diagram}, the outputs of the corresponding layers are passed through an addition layer, followed by a scale-and-shift layer to regulate the common outputs. Let $r_i$ and $g_i$ denote the outputs of the $i^{th}$ layer of ResNet-18 and GFNet, respectively. The combined output $\mathbf{z}_i$ of the $i^{th}$ layer is:
\begin{equation}
\mathbf{z}_i = a\cdot(\mathbf{r}_i+ \mathbf{g}_i) + b
\end{equation}
where $a$, $b$ are learnable parameters. The output $\mathbf{z}_i$ is passed to the next layer of both ResNet and GFNet. Finally, a $512$-dimensional feature vector is produced as the output for each image. This network is fine-tuned on the base task, where the ResNet-18 layers and scale-and-shift parameters are trainable, while the GFNet layers remain frozen during the fine-tuning process. The proposed loss function, as described in Sec. \ref{loss}, is utilized during this stage. A linear classification layer is added during fine-tuning and discarded afterwards. 

\subsection{Loss Functions} \label{loss}
We propose a loss function that combines focal  \cite{ross2017focal} and center loss \cite{wen2016discriminative}.

Focal loss prevents the large number of simple negative examples from confusing the detector, thereby focusing training on the more challenging examples:
\begin{equation}
L_F = -(1 - p_t)^\gamma \log(p_t)
\end{equation}
where $p_{t}$ is the predicted probability of the correct class and $\gamma$ is a non-negative tunable focusing parameter that adjusts the rate at which simple examples are down-weighted.

Center loss is employed to improve intra-class compactness, which can lead to more space in the feature space: 
\begin{equation}
L_C=\frac{1}{2} \sum_{i=1}^{m} \| \mathbf{F}_{i} - \mathbf{c}_{\mathbf{y}_{i}} \|_2^2.
\end{equation}
Here, $\mathbf{F}_{i}$ denotes the input features of $i^{th}$ class and $\mathbf{c}_{\mathbf{y}_{i}}$ denotes the centroid of features for $y_{i}^{th}$ class. 
The total loss is given by:
\begin{equation}
L = L_F + c \cdot L_C
\end{equation}
where $c$ is a hyperparameter set to $5\cdot 10^{-4}$. By combining these two loss functions, the model can learn representations with better discriminative power while improving generalization to unseen classes.

\section{Experimental Evaluation}

\subsection{Dataset and Experimental Settings}\label{data}

To evaluate the performance of the proposed framework and compare it to state-of-the art approaches in FSCIL SAR-ATR field, we employ the publicly available benchmark  MSTAR dataset \cite{keydel1996mstar}, which comprises SAR images of $10$ ground mobile targets. The proposed framework is evaluated against state-of-the-art FSCIL methods, including iCaRL \cite{rebuffi2017icarl},  CEC \cite{zhang2021few}, C-FSCIL \cite{hersche2022constrained}, FACT \cite{zhou2022forward}, S3C \cite{kalla2022s3c},  WaRP \cite{kim2023warping}, ODF \cite{kong2024few}, LUCIR \cite{hou2019learning}, TOPIC \cite{tao2020few}, IDLVQC \cite{chen2021incremental}, SAVC \cite{song2023learning}, CPL \cite{zhao2023few},  AASC \cite{zhao2024azimuth}, and DSSC \cite{zhao2024decoupled}.

The experiments were conducted in two setups, each with a specific class order configuration. Extensive experiments were performed across various scenarios to comprehensively assess the effectiveness of the proposed framework. The first setup, aligned with \cite{kong2024few} for a fair comparison, included the following experiments: $1$-way $5$-shot and $2$-way $5$-shot, where the base task comprises $4$ classes, and $1$-step $5$-shot, where the base task comprises 5 classes. An $n$-way $k$-shot scenario refers to a few-shot task involving $n$ classes, each with $k$ samples. The second setup followed \cite{zhao2023few, zhao2024azimuth, zhao2024decoupled} and the conducted experiments employed the $1$-way $5$-shot scenario, where the base task comprises $4$ classes. For the ablation study, the second setup was utilized. For a reliable assessment, all experiments were repeated $10$ times with different random seeds and the average of each metric is reported. For evaluation, standard FSCIL metrics are used: accuracy per incremental task $A_t$, average incremental accuracy $\bar{A}$, and performance drop,  $PD = A_0 - A_L$, where $A_L$ is the accuracy in the last task,  which quantifies catastrophic forgetting.

The proposed DILHyFS framework is trained for $30$ epochs using a learning rate of 0.01, employing stochastic gradient descent as the optimizer, a weight decay of $5 \cdot 10^{-4}$, and a batch size of $48$. The random projection layer has a dimension of $M = 10^{4}$. The focal loss function employs $\gamma = 0.5$. The training images were cropped to $88\times 88$, followed by random horizontal flipping.

\subsection{Results}\label{results}

To demonstrate the effectiveness of the proposed framework, comparisons were made against state-of-the-art FSCIL approaches for SAR-ATR using the MSTAR dataset. The results presented in Table~\ref{tab:resultMSTAR15}, highlight the superior performance of the DILHyFS framework compared to state-of-the-art methods in the $1$-way $5$-shot scenario. Notably, DILHyFS achieved the highest $\bar{A}$ attaining $84.06\%$, surpassing ODF \cite{kong2024few}, which yielded $82.47\%$. Furthermore, DILHyFS significantly outperformed all other methods in terms of $PD$, achieving  $10.94\%$, which represented a $68.64\%$ reduction compared to the ODF approach.

\begin{table}[htb]
    \caption{Comparison with state-of-the-art methods in the 1-way 5-shot scenario.}
    \label{tab:resultMSTAR15}
    \centering
    \resizebox{\columnwidth}{!}{%
        \begin{tabular}{lccccccccc}  
            \hline
            \multirow{2}{*}{Method} & \multicolumn{7}{c}{Accuracy in each task (\%)} & \multirow{2}{*}{$\bar{A}$ $\uparrow$}  & \multirow{2}{*}{PD $\downarrow$}\ \\
            \cline{2-8}
            & $0$ & $1$ & $2$ & $3$ & $4$& $5$& $6$ \\
            \hline
            iCaRL \cite{rebuffi2017icarl} & $99.45$ & $72.69$ & $77.66$ & $65.90$ & $58.48$ & $53.58$ & $52.09$ & $68.55$ & $47.36$  \\
            CEC \cite{zhang2021few} & $96.16$ & $85.99$ & $84.39$ & $78.05$ & $72.05$ & $62.89$& $58.50$& $76.86$& $37.66$ \\
            C-FSCIL \cite{hersche2022constrained} & $97.54$ & $86.20$ & $85.75$ & $76.60$ & $73.45$ & $64.40$ & $59.19$ & $77.59$ & $38.35$\\
            FACT \cite{zhou2022forward} & $99.44$ & $85.67$ & $85.68$ & $76.45$ & $73.30$ & $63.00$ & $57.65$& $77.31$& $41.79$\\
            S3C \cite{kalla2022s3c} & $\mathbf{99.70}$ & $\underline{89.35}$ & $\mathbf{92.46}$ & $80.19$ & $73.92$ & $68.41$ & $63.80$& $81.12$& $35.90$\\
            WaRP \cite{kim2023warping} &  $98.83$ & $88.24$ & $85.27$ & $77.26$& $73.51$& $64.02$   & $59.05$   & $78.03$ & $39.78$ \\
            ODF \cite{kong2024few} & $\underline{99.69}$ & $\mathbf{91.91}$ & $\underline{89.24}$ & $\underline{83.49}$ & $\mathbf{79.36}$ & $\underline{68.82}$ & $\underline{64.81}$& $\underline{82.47}$& $\underline{34.88}$\\
            \bf{DILHyFS (ours)} & $94.54$ & $80.21$ & $84.10$ & $\mathbf{85.08}$ & $\underline{78.98}$ & $\mathbf{81.93}$ & $\mathbf{83.60}$ & $\mathbf{84.06}$ & $\mathbf{10.94}$\\

            \hline
        \end{tabular}
    }
\end{table}

\begin{figure}[htbp]
\centering
\includegraphics[width=0.95\columnwidth]{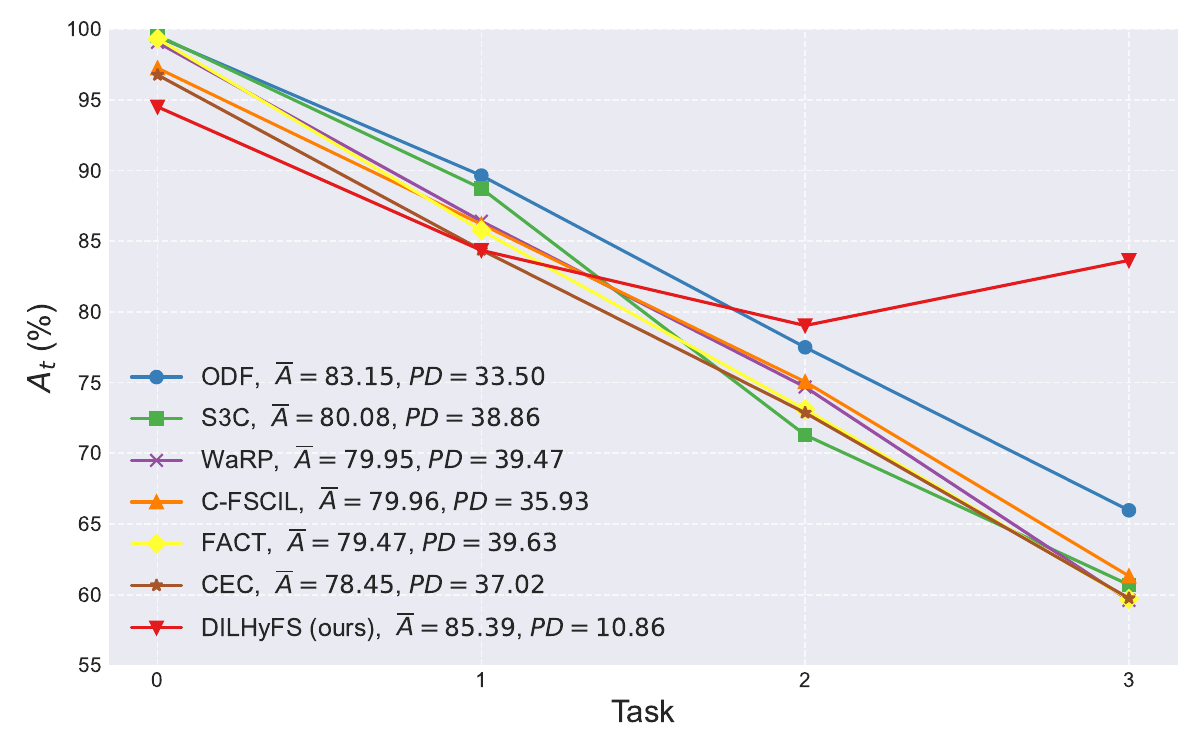}
\caption{Comparison with state-of-the-art methods in the 2-way 5-shot scenario.}
\label{fig:resultMSTAR2w}
\end{figure}

Fig.~\ref{fig:resultMSTAR2w} depicts the performance of the DILHyFS framework in the $2$-way $5$-shot scenario. DILHyFS achieved the highest $\bar{A}$ deriving $85.39\%$, outperforming the state-of-the-art ODF, which attained $83.15\%$. The third top-performing approach, namely S3C, achieved an $\bar{A}$ of $80.08\%$. Importantly, DILHyFS exhibited a remarkably low $PD$ of $10.86\%$, representing a significant  improvement of $67.6\%$ compared to ODF, which achieved $33.50\%$. These findings highlight the ability of DILHyFS to mitigate catastrophic forgetting while maintaining superior task performance in complex scenarios.

In the challenging 1-step 5-shot scenario, DILHyFS achieved the highest $\bar{A}$ reaching $89.59\%$, outperforming ODF, which yielded $84.17\%$. It is noteworthy that DILHyFS demonstrated a $PD$ of $1.89\%$, highlighting a remarkable $93.7\%$ reduction compared to ODF, which reached a $PD$ of $29.93\%$. Detailed results are shown in Table~\ref{tab:resultMSTAR1s5S}. These results emphasize the robustness of DILHyFS in such challenging scenarios that emulate real-word scenarios.

\begin{table}[htb]
    \caption{Comparison with state-of-the-art methods in the 1-step 5-shot scenario.}
    \label{tab:resultMSTAR1s5S}
    \centering
    \setlength{\tabcolsep}{4pt} 
    \renewcommand{\arraystretch}{1.2} 
    {\small 
        \resizebox{0.7\columnwidth}{!}{
            \begin{tabular}{lcccc}  
                \hline
                \multirow{2}{*}{Method} & \multicolumn{2}{c}{Accuracy in each task (\%)} & \multirow{2}{*}{$\bar{A}$ $\uparrow$}  & \multirow{2}{*}{PD $\downarrow$}\ \\
                \cline{2-3}
                & $0$ & $1$    \\
                \hline
                iCaRL \cite{rebuffi2017icarl} & $\underline{99.16}$ & $68.36$ & $83.76$ & $30.80$   \\
                
                
                CEC \cite{zhang2021few} & $96.68$ & $60.42$ & $78.55$ & $36.26$   \\

                C-FSCIL \cite{hersche2022constrained} & $95.79$ & $62.85$ & $79.32$ & $32.94$  \\
                
                FACT \cite{zhou2022forward} & $98.94$ & $64.07$ & $81.51$ & $34.87$   \\
                S3C \cite{kalla2022s3c} & $\mathbf{99.54}$ & $68.46$ & $84.00$ & $31.08$  \\
                
                WaRP \cite{kim2023warping} &  $98.72$ & $64.59$ & $81.66$ & $34.13$\\
                
                ODF \cite{kong2024few} & $99.13$ & $\underline{69.20}$ & $\underline{84.17}$ & $\underline{29.93}$   \\
                \bf{DILHyFS (ours)} & $90.54$ & $\mathbf{88.65}$ & $\mathbf{89.59}$ & $\mathbf{1.89}$   \\
    
                \hline
            \end{tabular}
        }
    }
\end{table}

Following the setup in state-of-the-art  \cite{zhao2024azimuth, zhao2024decoupled}, the proposed framework was compared against  state-of-the-art methods,  underlining its competitiveness and robustness in demanding real-world scenarios.  DILHyFS achieved the highest $\bar{A}$ attaining $91.14\%$ and outperformed DSSC \cite{zhao2024decoupled}, which yielded $81.94\%$. A significantly low $PD$ of $10.63\%$ was reported for the proposed framework, marking  an improvement of $66.32\%$ compared to state-of-the-art DSSC. These results demonstrate DILHyFS's effectiveness, as it outperforms state-of-the-art methods while maintaining high performance across various setups. Detailed results are shown in Table~\ref{tab:resultMSTARset2}.

\begin{table}[htb]
    \caption{Comparison with state-of-the-art methods in the 1-way 5-shot scenario for the second setup.}
    \label{tab:resultMSTARset2}
    \centering
    \resizebox{\columnwidth}{!}{%
        \begin{tabular}{lccccccccc}  
            \hline
            \multirow{2}{*}{Method} & \multicolumn{7}{c}{Accuracy in each task (\%)} & \multirow{2}{*}{$\bar{A}$ $\uparrow$}  & \multirow{2}{*}{PD $\downarrow$}\ \\
            \cline{2-8}
            & $0$ & $1$ & $2$ & $3$ & $4$& $5$& $6$ \\
            \hline
            iCaRL \cite{rebuffi2017icarl} & $92.12$ & $83.08$ & $73.46$ & $66.41$ & $65.04$ & $60.42$ & $54.27$ & $70.69$ & $37.85$  \\
            LUCIR \cite{hou2019learning} & $\mathbf{99.89}$ & $90.07$ & $76.97$ & $72.60$ & $70.33$ & $66.67$   & $60.89$& $76.67$& $39.00$\\ 
            TOPIC \cite{tao2020few} & $91.10$ & $85.27$ & $72.77$ & $63.25$ & $61.24$ & $56.03$  & $50.03$& $68.53$& $41.07$ \\
            IDLVQC \cite{chen2021incremental} & $97.02$ & $83.75$ & $71.12$ & $62.27$ & $59.54$ & $54.71$ & $49.20$& $68.23$& $47.82$\\
            CEC \cite{zhang2021few} & $90.54$ & $80.52$ & $72.27$ & $72.53$ & $66.97$ & $61.76$& $57.97$& $71.79$& $32.57$ \\
            FACT \cite{zhou2022forward} & $98.85$ & $87.36$ & $67.14$ & $64.24$ & $49.50$ & $46.69$ & $45.01$& $65.54$& $53.84$\\
            SAVC \cite{song2023learning} & $97.10$ & $89.43$ & $79.64$ & $75.66$ & $72.47$ & $68.48$ & $64.64$& $78.20$& $32.46$\\ \hline
            CPL \cite{zhao2023few} & $\mathbf{99.89}$ & $90.41$ & $78.32$ & $74.93$ & $73.04$ & $69.66$ & $63.66$& $78.56$& $36.23$\\
            AASC \cite{zhao2024azimuth} & $\mathbf{99.89}$ & $89.73$ & $76.33$ & $73.64$ & $71.46$ & $66.97$ & $62.31$& $77.19$& $37.58$\\
            DSSC \cite{zhao2024decoupled} & $\underline{99.47}$ & $\underline{93.84}$ & $\underline{83.01}$ & $\underline{80.25}$ & $\underline{76.60}$ & $\underline{72.48}$ & $\underline{67.91}$& $\underline{81.94}$& $\underline{31.56}$\\
            \bf{DILHyFS (ours)} & $94.63$ & $\mathbf{95.88}$ & $\mathbf{96.63}$ & $\mathbf{96.47}$ & $\mathbf{87.99}$ & $\mathbf{82.35}$ & $\mathbf{84.00}$ & $\mathbf{91.14}$ & $\mathbf{10.63}$\\
             
            \hline
        \end{tabular}
    }
\end{table}



\subsection{Cross-Domain Evaluation}

To further assess the robustness of the proposed framework, we conducted a cross-domain evaluation using MSTAR, OpenSARShip \cite{huang2017opensarship}, and SAR-Aircraft \cite{zhirui2023sar} datasets. The model first learns to classify military vehicles  and then incrementally adapts to ships and aircraft images. It is trained on $4$ base classes from MSTAR, with incremental tasks introducing $3$ new classes from MSTAR, followed by additional $3$ classes from each dataset. As shown in Table \ref{tab:cross}, DILHyFS  outperformed CEC and FACT across incremental tasks, achieving higher retention of previously learned classes while maintaining superior adaptation to new domains. We excluded DSSC, AASC, and CPL due to the unavailability of code, while SAVC was omitted as its performance was much lower than FACT and CEC.

\begin{table}[htb]
    \caption{Cross-domain evaluation of DILHyFS.}
    \label{tab:cross}
    \centering
    \resizebox{\columnwidth}{!}{%
        \begin{tabular}{lccccccc}  
            \hline
            \multirow{2}{*}{Method} & \multicolumn{5}{c}{Accuracy in each task (\%)} & \multirow{2}{*}{$\bar{A}$ $\uparrow$}  & \multirow{2}{*}{PD $\downarrow$}\ \\
            \cline{2-6}
            & $0$ & $1$ & $2$ & $3$ & $4$ \\
            \hline
            FACT \cite{zhou2022forward} & $\mathbf{99.38}$  & $70.69$ & $62.60$ & $59.93$ & $53.19$& $69.15$& $46.19$\\
            CEC \cite{zhang2021few}&  $\underline{96.31}$ & $\underline{75.60}$ & $\underline{65.79}$ & $\underline{60.93}$& $\underline{55.25}$& $\underline{70.77}$   & $\underline{41.06}$    \\ \hline
            \bf{DILHyFS (ours)} &  $94.83$ & $\mathbf{94.45}$ & $\mathbf{83.76}$ & $\mathbf{81.48}$ & $\mathbf{73.69}$ & $\mathbf{85.84}$ & $\mathbf{21.14}$\\

            \hline
        \end{tabular}
    }
\end{table}

\subsection{Ablation Study}

An ablation study was conducted to evaluate the individual performance of ResNet-18 and GFNet compared to the proposed DILHyFS framework. In the $1$-way $5$-shot setup, DILHyFS achieved an $\bar{A}$ of $91.14\%$, outperforming ResNet-18 and GFNet, which attained an accuracy of $88.80\%$ and $87.88\%$, respectively. Additionally, DILHyFS's $PD$ was $10.63\%$, which is comparable to ResNet-18 ($10.21\%$) and significantly lower than GFNet ($20.40\%$), demonstrating its ability to mitigate catastrophic forgetting. Detailed results are demonstrated in Table \ref{tab:ablation}.


\begin{table}[htb]
    \caption{Contribution analysis of the proposed framework backbones.}
    \label{tab:ablation}
    \centering
    \resizebox{\columnwidth}{!}{%
        \begin{tabular}{lccccccccc}  
            \hline
            \multirow{2}{*}{Method} & \multicolumn{7}{c}{Accuracy in each task (\%)} & \multirow{2}{*}{$\bar{A}$ $\uparrow$}  & \multirow{2}{*}{PD $\downarrow$}\ \\
            \cline{2-8}
            & $0$ & $1$ & $2$ & $3$ & $4$& $5$& $6$ \\
            \hline
            ResNet-18 &  $91.96$ & $93.61$ & $94.91$ & $95.19$& $86.50$& $77.69$   & $81.74$   & $88.80$ & $10.21$ \\
            GFNet  & ${96.11}$ & ${96.91}$ & ${96.69}$ & ${90.64}$ & ${82.57}$ & ${76.71}$ & ${75.70}$& ${87.88}$& $20.4$\\ \hline
            \bf{DILHyFS (ours)} & $94.63$ & $95.88$ & $96.63$ & ${96.47}$ & ${87.99}$ & ${82.35}$ & ${84.00}$ & ${91.14}$ & ${10.63}$\\

            \hline
        \end{tabular}
    }
\end{table}

\section{Discussion and limitations}

The proposed framework effectively mitigates catastrophic forgetting, achieving high accuracy across incremental tasks. However, it exhibits certain limitations. Despite outperforming state-of-the-art methods in average accuracy and performance drop, it struggles in the base task, where class separation is suboptimal, leading to lower accuracy. Additionally, the framework may face challenges when introducing new classes  from diverse datasets, as the learned feature space may not adapt optimally, hindering discrimination between base and novel classes.  Future work will aim at improving the performance of base class separation, which is expected to enhance accuracy across incremental tasks.

\bibliographystyle{IEEEbib}
\bibliography{bibliography}

\end{document}